\documentclass[runningheads]{llncs}
\usepackage{graphicx}
\usepackage{comment}
\usepackage{amsmath,amssymb}
\usepackage{color}

\begin{document}
\pagestyle{headings}
\mainmatter

\title{Consistent Multiple Sequence Decoding} 
\titlerunning{Consistent Multiple Sequence Decoding}

\author{Bicheng Xu\inst{1,2} \and Leonid Sigal\inst{1,2,3}}
\authorrunning{B. Xu \and L. Sigal}

\institute{The University of British Columbia, Vancouver, Canada \and Vector Institute for AI, Toronto, Canada \and Canada CIFAR AI Chair \\
\email{\{bichengx, lsigal\}@cs.ubc.ca}}

\maketitle

\begin{abstract}
Sequence decoding is one of the core components of most visual-lingual models. However, typical neural decoders when faced with decoding multiple, possibly correlated, sequences of tokens resort to simple independent decoding schemes. In this paper, we introduce a consistent multiple sequence decoding architecture, which is while relatively simple, is general and allows for consistent and simultaneous decoding of an arbitrary number of sequences. Our formulation utilizes a {\em consistency fusion mechanism}, implemented using message passing in a Graph Neural Network (GNN), to aggregate context from related decoders. This context is then utilized as a secondary input, in addition to previously generated output, to make a prediction at a given step of decoding. Self-attention, in the GNN, is used to modulate the fusion mechanism locally at each node and each step in the decoding process. We show the efficacy of our consistent multiple sequence decoder on the task of dense relational image captioning and illustrate state-of-the-art performance (+ 5.2\% in mAP) on the task. More importantly, we illustrate that the decoded sentences, for the same regions, are more consistent (improvement of 9.5\%), while across images and regions maintain diversity.

% Consistency is a commonly existed correlation among sequences. When decoding multiple sequences at the same time, current existing works all assume that every sequence is independent. That is, each sequence is independently decoded. However, many times in real life, many sequences are not independent with each other. That is, there exist correlations, and commonly, consistent patterns among them. In this work, we propose a consistent multiple sequence decoder with a novel consistency fusion mechanism to incorporate consistency during decoding. Our consistent decoder do provide better and more consistent multiple sequence decoding results than independent decoders. \red{Revisise.}

\end{abstract}

\section{Introduction}
Sequence decoding has emerged as one of the fundamental building blocks for a large variety of computer vision problems. For example, it is a critical component in a range of visual-lingual architectures, for tasks such as image captioning~\cite{li2019entangled,olivastri2019end,shen2017weakly,vered2019joint} and question answering~\cite{anderson2018bottom,antol2015vqa,lu2016hierarchical}, as well as in generative models that tackle trajectory prediction or forecasting~\cite{alahi2016social,kosaraju2019social,lee2017desire,tang2019multiple}. Most existing methods assume a single sequence and implement neural decoding using recurrent architectures, e.g., LSTMs or GRUs; recent variants include models like BERT~\cite{devlin2018bert}. However, in many scenarios, more than one sequence needs to be decoded at the same time. Common examples include trajectory forecasting in team sports~\cite{felsen2017will,sun2019stochastic,yeh2019diverse} or autonomous driving~\cite{park2018sequence,rhinehart2019precog}, where multiple agents (players/cars) need to be predicted and behavior of one agent may closely depend on the others. 

Similarly, in dense image  captioning~\cite{johnson2016densecap,yang2017dense,yin2019context}, or dense relational captioning~\cite{kim2019dense}, multiple linguistic captions need to be decoded for a set of possibly overlapping regions, with the goal of describing the regions themselves~\cite{johnson2016densecap} or relationships among them~\cite{kim2019dense}. In such scenarios, a trivial (and most common) solution is to decode each sequence independent of the other. While simple, this looses the important context among the various sequences being processed.

Consider, for example, dense image relational captioning, which is the main focus of this paper. As defined in~\cite{kim2019dense}, the goal of the task is to generate captions denoting relationships between pairs of objects in an image. This means that, in general, multiple captions will refer to the same object region as either the {\em subject} or {\em object} of the relation that needs to be captioned. If processed independently, each decoder will interpret the same image region and combine it with a language model to produce a caption. This may result in different proper names (possibly synonyms or worse) being used in each decoder, e.g., “\underline{desk} (subject) standing on the floor”, “lamp standing on the \underline{furniture} (object)”, “\underline{table} (subject) is next to a chair” when referring to the exact same image region containing a table. This seems unnatural and problematic, as a given person typically would have a preferred expression for an object and would use it consistently; while different people may use different terms or names to refer to a given object, any one individual would typically be self-consistent and not switch naming convention. In other words, a given individual would use “table” or “desk” in all three aforementioned captions, but not switch between these phrases. A more drastic example is when one sequence is directly correlated with another (e.g., a player following an opponent; prediction of temperature in close geographic regions).    

Addressing this challenge in the multi-sequence decoder, implies coupling various decoders so they can communicate and consistently generate sequences. %(e.g., references in dense relational captioning). 
% This problem of consistent multi-sequence decoding is related to multi-agent state or trajectory prediction, which has been handled by contextual spatial pooling in SocialLSTM~\cite{alahi2016social} or graph-structured variational recurrent neural network in~\cite{sun2019stochastic}. Different from these works, which are designed to operate specifically on 2D trajectory data, we propose a more flexible and effective multi-sequence consistent decoding architecture.
We propose a model, where specifically, at each time step during decoding, all decoders are allowed to communicate among themselves using a graph neural network to arrive at the context required to generate the next output (word or 2D location). We note that depending on the application, the communication may take the form of propagating the hidden state or the readout (output) from each decoder, which may involve sampling (choosing a word). We illustrate the performance of this architecture on both synthetic data and real application of dense relational captioning, where we simultaneously generate each word, in each sequence, conditioned on the previous words of all sequences. We illustrate improvements in the quality of the captions and, more significantly, consistency.

\vspace{0.1in}
\noindent
Our {\bf contributions} are as follows:
\vspace{-0.05in}
\begin{itemize}
  \item We propose a novel consistent multiple sequence decoder which can learn to incorporate consistent patterns among sequences, while decoding them. 
  \item The consistency fusion mechanism in the proposed decoder, implemented using message passing in a graph neural network, enables arbitrary topology of problem-dependent contextual information flow.
  % provides useful information when decoding sequences that exhibit correlation  patterns.  
  \item We validate our approach by producing new state-of-the-art results on the dense relational captioning task, resulting in $5.2\%$ improvement in mAP measurement, obtained using consistent decoding mechanism alone.
  \item We introduce a new measurement to evaluate the consistency in decoded captions of the dense relational captioning task, and illustrate $9.5\%$ improvement on this measure. %We also introduce a new measurement to evaluate the consistency in decoded sequences (captions), and illustrate $9.5\%$ improvement on this measure. 
  % in a real world experiment.
  % \item We validate our approach by producing new state-of-the
\end{itemize}
We note that while our main focus is on dense image relational captioning, % as our main application domain, 
the formulation of our consistent multiple sequence decoder is general and can be used in any application and architecture which requires multi-sequence decoding.

\section{Related Work}

\subsection{Multiple Sequence Decoding}
Sequence decoding is about giving some condition information, such as images or videos, and decoding sequences from the given condition. Typical applications are image/video captioning~\cite{huang2019adaptively,olivastri2019end,rennie2017self,anderson2018bottom,pan2017video,gao2017video}, and dense image/video captioning~\cite{johnson2016densecap,yang2017dense,yin2019context,krishna2017dense,li2018jointly,rahman2019watch}. Image/video captioning takes an image or a short video clip as an input and automatically produces a caption describing the content. Dense captioning, instead of generating one caption for the whole image or video, is tasked with generating a caption for every region of interest (for image), or every time period of interest (for video). 

% is given an image or a short video clip, we want a model to generate a caption describing the content for us automatically. Dense captioning is instead of generating one caption for the whole context, we want a model to generate a caption for every region of interest (for image), or every time period of interest (for video). Image/video captioning only decodes one sequence given the context, while dense captioning decodes multiple sequences at the same time. 

The task of dense image captioning was first proposed in~\cite{johnson2016densecap}, where given an image, the model is able to detect the objects in that image, and then generate a caption for every detected object. Recently, Kim et al.~\cite{kim2019dense} extended the task to generate a caption for every pair of detected objects, which is called dense relational captioning. These two tasks are both multiple sequence decoding tasks. However, all these works generate captions independently of one another, and do not consider the potential correlations existed among sequences to be decoded.
Similar to Kim et al.~\cite{kim2019dense}, we address the dense relational captioning task, but specifically focus on modeling correlations among language decoders using our proposed consistent multiple sequence decoding architecture.
%  , even for the dense relational captioning task where there exists correlation among sequences.  

\subsection{Multiple Sequence Prediction}
Sequence prediction is about giving some state information in the past, predicting the future states. 
%The tasks range from trajectory prediction~\cite{alahi2016social,kosaraju2019social,lee2017desire} to future world state prediction~\cite{sun2019stochastic,liu2019naomi}. 
The problem of consistent multi-sequence decoding is related to multi-agent state or trajectory prediction, which has been handled by contextual spatial pooling in SocialLSTM~\cite{alahi2016social} or graph-structured variational recurrent neural network in~\cite{sun2019stochastic}. SocialLSTM~\cite{alahi2016social} predicts the future trajectories of people based on their past positions. It proposes the idea of social pooling to incorporate a person's neighbors' past position information for one person's trajectory prediction. Sun et al.~\cite{sun2019stochastic} proposes a graph-structured variational recurrent neural network model that learns to integrate temporal information with ambiguous visual information to forecast the future states of the world, which is represented as multiple sequences. This model works specifically on sports videos. \cite{liu2019naomi} proposes a non-autoregressive decoding procedure for deep generative models that can impute missing values for spatiotemporal sequences with long-range dependencies, which can also be applied to forward prediction.

Different from these works, which are designed to operate specifically on 2D trajectory data, we propose a more flexible and effective multi-sequence consistent decoding architecture.  

% \red{Leon: Need section on GNN as well us beef up above.}
% \subsection{Graph Neural Network}

\section{Consistent Multiple Sequence Decoder}
\begin{figure}[t]
  \centering
  \includegraphics[width=\textwidth]{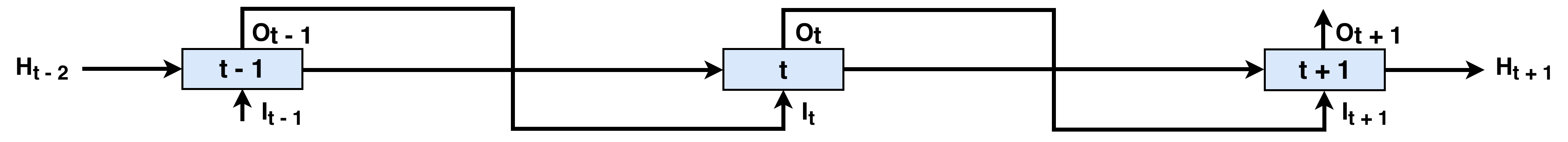}
  \caption{{\bf Independent Decoder.} In the illustration, $H$, $I$, and $O$ refer to the hidden state, input, and output representations at time steps $t - 2$, $t - 1$, $t$, and $t + 1$ respectively. In the independent decoder, the input at the current time step is the output representation from its own previous time step.}
  \label{fig:model_independent}
\end{figure}

Sequence decoding is usually performed by recurrent neural networks (RNNs). Traditionally, multiple sequence decoding decodes every sequence independently. %  An independent decoder is shown in Figure~\ref{fig:model_independent}. 
During decoding, at every time step, the input to the recurrent unit is only the output representation from its own previous time step (see Figure~\ref{fig:model_independent}), or the representation of a special token (e.g., the {\tt start} or {\tt end} of sequence token).

However, in general, the sequences may not be independent of one another and may exhibit high degree of correlation. In such cases, the independent decoder will loose out on the across-sequence context and correlations, making predictions poor or inconsistent. Therefore it is important to learn problem-dependent correlations among the decoders and to be able to integrate this contextual information effectively and continuously as the decoding takes place. 

% in real life, when decoding multiple sequences simultaneously, these sequences to be decoded are not always independent with each other. That is, some sequences may be correlated with each other. In this case, the independent decoder can not utilize the existed correlations for better decoding results. There can be many correlations existing among multiple sequences, and one common correlation is consistency. That is, there may exist some common patterns among sequences.  

\begin{figure}[t]
  \centering
  \includegraphics[width=\textwidth]{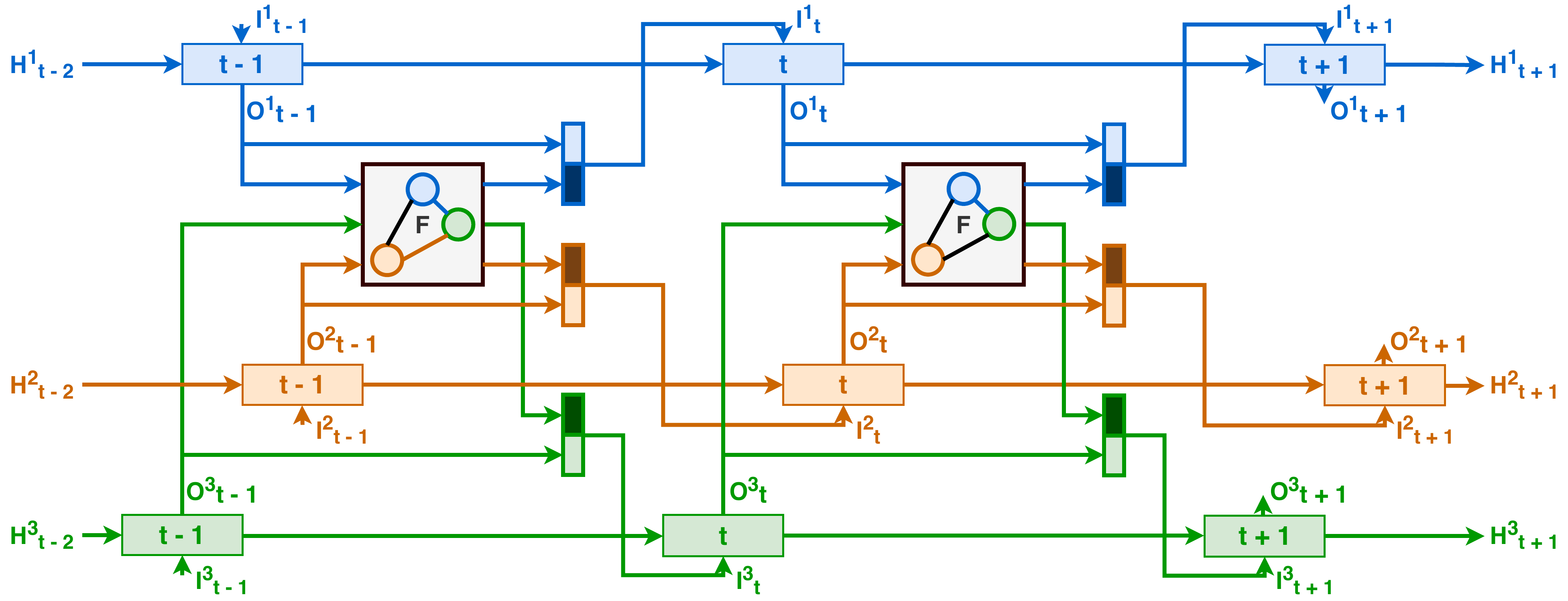}
  \caption{{\bf Consistent Multiple Sequence Decoder.} Illustrated is an example of decoding three sequences simultaneously, indicated by blue, orange, and green colors. The $H$, $I$, and $O$ refer to the hidden state, input, and output representations of the relevant recurrent unit at time steps $t - 2$, $t - 1$, $t$, and $t + 1$ respectively. $\operatorname{F}$ represents the consistency fusion mechanism, which acts on the output representations $O$ from all decoders. At every time step, our decoder receives two inputs. One is the output representation generated by itself from the previous time step; the other is fused output representation from $\operatorname{F}$. The two inputs are concatenated and fed into a recurrent unit.}
  \label{fig:model_decoder}
\end{figure}

Our proposed consistent multiple sequence decoder, consistent decoder in short, can learn and take advantage of the consistency/correlation pattern existing among sequences for better decoding results. To utilize these consistency patterns, we want each decoder to have the summary of the previous outputs from other decoders when generating the output. Therefore, during decoding, at every time step, rather than receiving one input, our consistent decoder gets two. First is the output representation from the previous time step of its own, as in the traditional independent decoder. The second input is a fused output representation from previous time step of all other sequence decoders that it may be correlated with. We propose a novel graph-based consistency fusion mechanism (Section~\ref{sec:fusion}) for this output fusion. The proposed fusion mechanism has two core benefits: (1) it allows modeling of arbitrary dependencies among the decoders (i.e., it allows one to define chains of potential directed or undirected influences among the decoders, e.g., {\tt Decoder A} can influence {\tt Decoder B} and {\tt C} and vice versa, {\tt Decoder C} can influence {\tt Decoder D} but {\em not} vice versa); and (2) a graph attention mechanism can guide contribution of information at each decoding step, allowing the context to be dynamic over time. The overall structure of our consistent multiple sequence decoder is illustrated in Figure~\ref{fig:model_decoder}. 

\subsection{Consistency Fusion Mechanism} \label{sec:fusion}
% \red{Leon: This section needs to be revised.}
Graph neural networks (GNNs)~\cite{kipf2016semi,knyazev2019understanding,velivckovic2017graph,xu2018powerful} are powerful mechanisms for contextual feature refinement. As such, we use gated GNN~\cite{li2015gated}, a type of graph neural networks, to implement the consistency fusion mechanism. In doing so, our goal is to propagate information among related decoders in order to arrive at a {\em fused} representation for each decoding step in each of the decoders.

% in feature representation learning~\cite{kipf2016semi,xu2018powerful,velivckovic2017graph,knyazev2019understanding}. % especially when one feature will be effected by some other features \red{Need some references}. 
% In our consistent decoder, we want one of the inputs, the fused output representation, to be a powerful feature representation of the outputs from all decoders. We use gated graph neural network~\cite{li2015gated}, a type of graph neural networks, to implement the consistency fusion mechanism. % The aim is to capture the consistency among multiple sequences, and to provide useful information during decoding. 

During each step of decoding, this mechanism fuses the output representations of the previous time step from all decoders. A graph is built on these output representations. Every output representation is treated as a node in the graph and the number of nodes is equal to the number of sequences/decoders. Formally, $\mathcal{G} = (\mathcal{V}, \mathcal{E})$, where $\mathcal{V}$ is the set of vertices in a graph, % . The vertices are associated with hidden feature representations consisting of encoded outputs from all decoders at one time step, 
and $\mathcal{E}$ is the adjacency matrix representing the dependencies between any pair of nodes/decoders ($\mathcal{E} \in \{0,1\}^{|\mathcal{V}| \times |\mathcal{V}|}$). Message passing is used to refine these representations and the results are fed into corresponding decoders as additional context. 
% Note that this process proceeds for each step in the decoding. 

\vspace{0.1in}
\noindent
{\bf Adjacency matrix.}
The adjacency matrix in the gated GNN is predefined before sequence decoding and same adjacency matrix is used at every time step. We say that two sequences are correlated, if they are believed to be somehow dependent. We set the elements in the adjacency matrix $\mathcal{E}(i,j)=1$ if two sequences $i$ and $j$ are correlated, and $\mathcal{E}(i,j)=0$ otherwise. Note that reasonable adjacency matrix can often be estimated based on the constraints of the problem; e.g., for dense relational captioning all decoders that take a certain region into account (or regions with sufficient IoU overlap) can be connected. In problem where no prior information is known, a complete adjacency matrix can be used. 

%All sequences are decoded at the same time.
% Then based on the defined adjacency matrix, we build our graph neural network.

\vspace{0.1in}
\noindent
{\bf Gated graph neural network.}
We use gated GNN~\cite{li2015gated} to fuse the output representations at every time step based on the predefined adjacency matrix. This involves two steps in one iteration of feature update. The first step is $\operatorname{AGGREGATE}$ (message passing), and the second step is $\operatorname{COMBINE}$ (feature fusion). The $\operatorname{AGGREGATE}$ step can be formulated as follows:
\begin{equation}
\label{eq:aggregate}
\begin{aligned}
    \mathbf{a}_v^{(k)} &= \operatorname{AGGREGATE}(\{\mathbf{h}_u^{(k-1)}: u \in N(v)\}) \\
    &= \sum_{u \in N(v)} \alpha_u^{(k)} \left( \mathbf{W} \cdot \mathbf{h}_u^{(k-1)} \right),
\end{aligned}
\end{equation}
where $\mathbf{a}_v^{(k)}$ is the aggregated information received by node $v$ from all its neighbors $N(v)$ during $k^{th}$ iteration of message passing. $\mathbf{h}_u^{(k-1)}$ is a $d$-dimensional vector representing the feature of node $u$ before the message passing. $\mathbf{W}$ is a learnable $d \times d$-dimensional graph kernel matrix and $\alpha_u^{(k)}$ is the adaptive self-attention which will be described later.

After obtaining $\mathbf{a}_v^{(k)}$, the feature of node $v$ will be updated using the $\operatorname{COMBINE}$ operation,
\begin{equation}
    \mathbf{h}_v^{(k)} = \operatorname{COMBINE}(\{\mathbf{h}_v^{(k-1)}, \mathbf{a}_v^{(k)}\}).
\end{equation}
We adopt the GRU gating mechanism~\cite{cho2014properties} in the $\operatorname{COMBINE}$ operation as proposed in~\cite{li2015gated}. After $K$ iterations of $\operatorname{AGGREGATE}$ and $\operatorname{COMBINE}$, we get $\mathbf{h}_v^{(K)}$, the final representation of node $v$, which is our fused output representation. 

\vspace{0.1in}
\noindent
{\bf Initialization.}
The initial state of each node $\mathbf{h}_v^{(0)}$ is set to the output representation produced by the recurrent unit of the corresponding decoder.   

\vspace{0.1in}
\noindent
{\bf Adaptive self-attention.} 
We calculate the self-attention $\alpha_u^{(k)}$ for every node $u$ at each message passing iteration $k$ to decide the edge weights of the graph. For efficiency, all edges that originate from a node $u$ have the same weight $\alpha_u^{(k)}$, which is calculated by
\begin{equation}
    \alpha_u^{(k)} = \operatorname{fc_{att}}(\mathbf{W} \cdot \mathbf{h}_u^{(k-1)}),
\end{equation}
where $\mathbf{W} \cdot \mathbf{h}_u^{(k-1)}$ is the same term as in Equation~\ref{eq:aggregate}, and $\operatorname{fc_{att}}(\cdot)$ is a three-layer fully-connected neural network with leaky ReLU activations~\cite{maas2013rectifier}. When aggregating the information for a node $v$, for all of its neighbors $u$, $\alpha_u^{(k)}$ is normalized to sum to 1 by the $\operatorname{softmax}$ operation.

We do not calculate a pair-wise attention between all connected nodes as it is computationally expensive to do so for a large graph. The number of elements in the attention matrix increases quadratically with the increasing number of nodes. As can be seen in the dense relational captioning experiments (Section~\ref{sec:relational_captioning}), the average number of sequences to decode for every image is $209$ during evaluation; making pair-wise attention untenable. 

\subsection{Input Combination}
At every time step, our consistent decoder receives two inputs. One is the output representation $\mathbf{r}_{own}$ from its own previous time step, and the other is the fused output context representation $\mathbf{r}_{fuse}$ from the consistency fusion mechanism. We simply concatenate $[\mathbf{r}_{fuse}, \mathbf{r}_{own}]$ to form the input to the recurrent unit. %  at every time step. 
%We use a special start of sequence token to start the sequence decoding. 

% \red{decoding distribution ?}

\section{Experiments}

We conduct two sets of experiments. One is on a synthetic mathematical dataset created by us, where we build two mathematical sequences one being a function of the other. The goal is to illustrate the full power of our model in the scenario where dependency between sequences is strong; as well as to illustrate the power of consistent decoder in absence of additional complexities introduced by a specific vision problem. The second set of experiments focuses on the use of consistent multiple sequence decoding for dense relational captioning on a real image dataset introduced in~\cite{kim2019dense}. We also conduct ablation experiments to illustrate the importance of various components.  

\subsection{Mathematical Sequence Decoding} \label{sec:math_decoding}
\subsubsection{Dataset.} We build a dataset which contains of paired mathematical sequences: 
\begin{align}
    y_1(x) &= a \cdot x + b; \\ 
    y_2(x) &= c \cdot x + d + y_1.
\end{align}
By design, $y_2(x)$ is a function of $y_1(x)$ and hence depends on the value of $y_1(x)$. The constants $a$, $b$, $c$, and $d$ are randomly drawn from the uniform distribution $\mathcal{U}(5, 15)$. Paired sequences are constructed by evaluating the two functions on integer values $x \in [1, 16]$.
The goal of the task is to forcast values of both $\mathbf{y}_1 = [y_1(2), ..., y_1(16)]$ and $\mathbf{y}_2 = [y_2(2), ..., y_2(16)]$ for $15$ steps conditioned on the input $\mathbf{x}_1 = [a, b, y_1(1)]$ and $\mathbf{x}_2 = [c, d, y_2(1)]$ respectively. 
% During decoding, we give the model the values of $a$, $b$, $c$, $d$, and the values of $y_1$ and $y_2$ evaluated at $x = 1$. We let the model generate the remaining $15$ values in the sequence. 
We build 80K paired sequences in total; 70K for training and 5K for validation/testing each.  

\begin{figure}[t]
  \centering
  \includegraphics[trim=0 0.14in 0 0, clip, width=\textwidth]{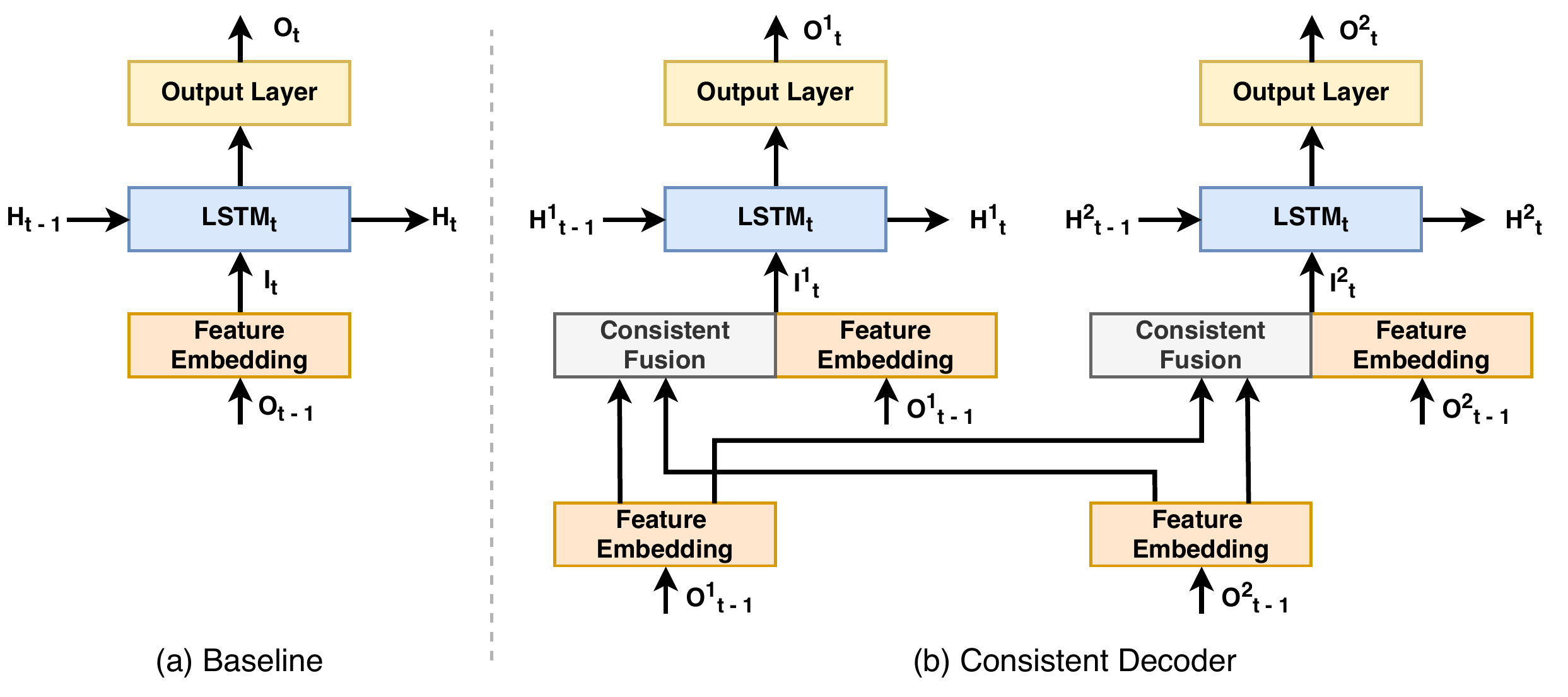}
  \caption{{\bf Synthetic Experiment.} One recurrent unit at time step $t$ of the models. (a) The baseline model, where the LSTM input is its own output representation from the previous time step. (b) Our consistent decoder, where the LSTM unit receives two inputs. One is its own output representation from the previous time step, the other one is the output from the consistent fusion mechanism. We show the two decoders here. The consistent fusion mechanism is applied on the two decoders' output representations.}
  % \vspace{-0.2in}
  \label{fig:model_math}
\end{figure}

% \vspace{0.1in}
% \noindent
% {\bf Models.} We build two baselines, named baseline and extended baseline below, and one consistent decoder model on this dataset.

\vspace{0.1in}
\noindent
\textbf{Baseline.} As a baseline, we train two separate decoders to decode the sequences independent of one another, where every decoder is an LSTM~\cite{hochreiter1997long} recurrent neural network. One such recurrent unit is illustrated in Figure~\ref{fig:model_math}, which contains three components: a feature embedding layer, an LSTM unit, and an output layer. In the first time step, the input to the feature embedding layer is a concatenation of $a$ and $b$, or $c$ and $d$. The input in the second time step is $y_1(1)$ or $y_2(1)$ value evaluated at $x = 1$. The output layer is trained to produce the next value in the sequence. The input in the remaining time steps is the output from the previous time step. Except the feature embedding layer in the first time step, all other feature embedding layers share the same weights in one decoder. All LSTM units and output layers have the same weights in one decoder. 

\vspace{0.1in}
\noindent
\textbf{Baseline (2x).} Our consistent decoder takes two inputs, increasing the input dimension of the LSTM cell by a factor of two (compared to the baseline above). To ensure that the improved performance is not due to the increased capacity of the LSTM, we also implement a baseline where input to LSTM cell matches ours; we call this variant Baseline (2x). In practice, we simply concatenate the two identical inputs and feed this vector into LSTM unit instead. We highlight that the number of parameters in the LSTM of this variant matches exactly those in our consistent decoder. 

% To mimic the structure of the consistent decoder, we make another baseline which also receives two inputs. These two inputs are concatenated to feed into the LSTM unit. In the baseline setting, during decoding, one decoder does not know the output from the other decoder. Built upon the baseline, besides one input as the output from the previous time step, we set the other input same as this one \red{(so weird, how to revise?)}. One such recurrent unit is shown in \red{Figure~\ref{fig:model_mathbase2}}.

\vspace{0.1in}
\noindent
\textbf{Consistent decoder.} Our consistent decoder (see Figure~\ref{fig:model_math}) has the same structure as the Baseline (2x) above. The difference is that we run the consistent fusion mechanism, that involves message passing in a gated GNN, to obtain the fused context vector that is concatenated to the encoding of the previous output. 
% , while we apply the consistency fusion mechanism onto the output embedding \red{not clear enough}. The other input to the LSTM unit is the output from the consistency fusion mechanism, a fused representation of the two outputs from the two decoders. 
Because only two sequences are decoded in this scenario, we set the number of $\operatorname{AGGREGATE}$ and $\operatorname{COMBINE}$ iteration ($K$) in the fusion mechanism to $1$. 
% One such recurrent unit is shown in Figure~\ref{fig:model_math}. 

\vspace{0.1in}
\noindent
\textbf{Implementation details.} The LSTM input size is set to 32 for Baseline; and 64 for both Baseline (2x) and our Consistent Decoder. The hidden state size is set to 2048. The total steps for the LSTM decoder is $16$, where the first time step is encoding coefficients ([$a$, $b$] for $y_1$ or [$c$, $d$] for $y_2$); and the remaining ones are used for sequence decoding. The second input in the first two time steps of the Baseline (2x) and the Consistent Decoder is a zero vector. % \red{add specific network structure?} 
During training, we set batch size to $40$. 
% , which means that every decoder decodes $40$ sequences at the same time. % \red{do we really need the second half?}. 
We use SGD optimizer with momentum $0.98$, and a constant learning rate $10^{-6}$. Models are trained using prediction mean squared error (MSE) loss. % We train the model for 500 epochs and pick the model based on the best validation loss.  
% The performance is similarly measured by MSE.

\vspace{0.1in}
\noindent
\textbf{Results.} The quantitative results are shown in Table~\ref{table:math_quan}, which are measured using the MSE loss on the test set for $y_1$ and $y_2$ sequences respectively. All three models can do well in decoding $y_1$. However, baseline models are not able to do well when decoding $y_2$. Our consistent decoder, learns to incorporate predictions from $y_1$ when decoding $y_2$, resulting in a much more accurate decoding. The decoding for $y_2$, using consistent decoder, has $1,000$ times lower MSE loss than the baselines! 
% every decoder knows the output from the other decoder when generating outputs, it can decode $y_2$ \red{accurately?}. 
% Further, due to the consistency fusion mechanism, the consistent decoder achieves lower loss on $y_1$ as well, decoding/forcasting $y_1$ nearly perfectly. 
One qualitative result is shown in Table~\ref{table:math_qual} with $a = 14.56$, $b = 5.18$, $c = 10.93$, and $d = 14.66$ for decoding $y_2$. 
%\red{Add more observations and discussions!}

\begin{table}[t]
\setlength{\tabcolsep}{3pt}
\begin{center}
\caption{{\bf Synthetic Experiment.} Results on the mathematical sequence dataset.}
\label{table:math_quan}
\begin{tabular}{l|c|c}
\hline
Model & MSE loss on $y_1$ & MSE loss on $y_2$ \\
\hline
\hline
Baseline & 0.0200 & 340.1852 \\
Baseline (2x) & 0.0043 & 340.0925 \\
Consistent Decoder & 0.0049 & {\bf 0.3356} \\
\hline
\end{tabular}
\end{center}
\vspace{-0.15in}
\end{table}

\begin{table}[t]
\setlength{\tabcolsep}{3pt}
\begin{center}
\caption{{\bf Synthetic Experiment.} A qualitative result on the mathematical sequence dataset with $a = 14.56$, $b = 5.18$, $c = 10.93$, and $d = 14.66$ for decoding $y_2$.}
\label{table:math_qual}
\begin{tabular}{l|cccccccc}
\hline
Time step & ($x = 1$) & 2 & 3 & 4 & 5 & 6 & 7 & 8 \\
\hline
Baseline & 45.33 & 66.37 & 87.19 & 107.52 & 128.24 & 148.93 & 169.57 & 190.28 \\
Baseline (2x) & 45.33 & 66.63 & 86.99 & 107.37 & 128.08 & 148.58 & 169.14 & 189.73 \\
Consistent decoder & 45.33 & 66.72 & 95.09 & 121.88 & 147.06 & 172.58 & 198.09 & 223.54 \\
\hline
Ground-truth & 45.33 & 70.83 & 96.32 & 121.81 & 147.31 & 172.80 & 198.30 & 223.79 \\
\hline
\hline
Time step & 9 & 10 & 11 & 12 & 13 & 14 & 15 & 16 \\
\hline
Baseline & 210.82 & 231.54 & 252.40 & 273.14 & 293.95 & 314.66 & 335.11 & 355.56 \\
Baseline (2x) & 210.40 & 231.24 & 252.24 & 273.37 & 294.50 & 315.54 & 336.20 & 356.61 \\
Consistent decoder & 248.99 & 274.44 & 299.89 & 325.35 & 350.80 & 376.25 & 401.76 & 427.31 \\
\hline
Ground-truth & 249.28 & 274.78 & 300.27 & 325.76 & 351.26 & 376.75 & 402.24 & 427.74 \\
\hline
\end{tabular}
\end{center}
\vspace{-0.15in}
\end{table}

\subsection{Dense Relational Captioning} \label{sec:relational_captioning}
The task of dense relational captioning was first introduced in~\cite{kim2019dense}. It is an extension of the dense image captioning initially proposed by Johnson et al. in~\cite{johnson2016densecap}. The dense relational captioning task is similar to pipeline in~\cite{johnson2016densecap}, where regions are detected first and then captioned. The main difference is that relational captions are generated for every {\em pair} of detected object regions.

% It follows the detection and then captioning pipeline as in~\cite{johnson2016densecap}. The task is about given an image, detecting the objects in the image first, and then generating a caption for every pair of detected objects. 

\vspace{0.1in}
\noindent
\textbf{Dataset.}
Dense relational captioning dataset, introduced in~\cite{kim2019dense}, is built upon the Visual Genome~\cite{krishna2017visual} (VG, v.1.2). It consists of 75,456 training/4,871 validation/4,873 testing images. The ground-truth relational captions are formed using the relationship labels and attribute labels of VG. We use original splits proposed in~\cite{krishna2017visual} for our experiments. However, we also find an unusual lack of consistency in the dataset among the captions for the same object. This stems from merging of different annotations provided by Amazon Mechanical Turkers for a given image in VG. Because we expect captions produced by one person to be (self-)consistent, we also report the performance on a manually constructed variant of the dense relational captioning dataset which has consistent labels. % with these properties.
%and captions produced across population of turkers to be diverse, 
%we also report the performance on manually constructed variants of the dense relational captioning dataset with these properties.  

% However, we find that in the original dense relational captioning labels, the descriptions for one bounding box (object) may not be consistent. That is, there may exist multiple descriptions for one specific bounding box in different relations in an image. \red{better to have some statistics.} \red{To show the effectiveness of our consistent decoder at most?}, we build another consistent caption label set, called consistent label set, where every bounding box (object) only has one ground-truth description. The details of the two ground-truth label sets are below.

\vspace{0.1in}
\noindent
\textbf{Original label set.} This is the original dense image relational captioning label set, which has vocabulary size of 15,596 words across all captions.

\vspace{0.1in}
\noindent
\textbf{Consistent label set.} For every image, we first count the number of different descriptions for every bounding box. Then for the bounding box which has multiple descriptions, we select the most frequent description as its only ground-truth. This makes the descriptions for one bounding box consistent for a specific image. After label pre-process~\cite{kim2019dense}, we obtain a vocabulary size of 15,009 words. %on this label set.

% \noindent \textbf{Diverse label set.?}
% \noindent \textbf{Sampled Labels.} For every image, we randomly sample a round-truth description for every bounding box, but in one image, the descriptions for one bounding box is consistent. The vocabulary size is 15114.

\begin{figure}[t]
  \centering
  \includegraphics[width=\textwidth]{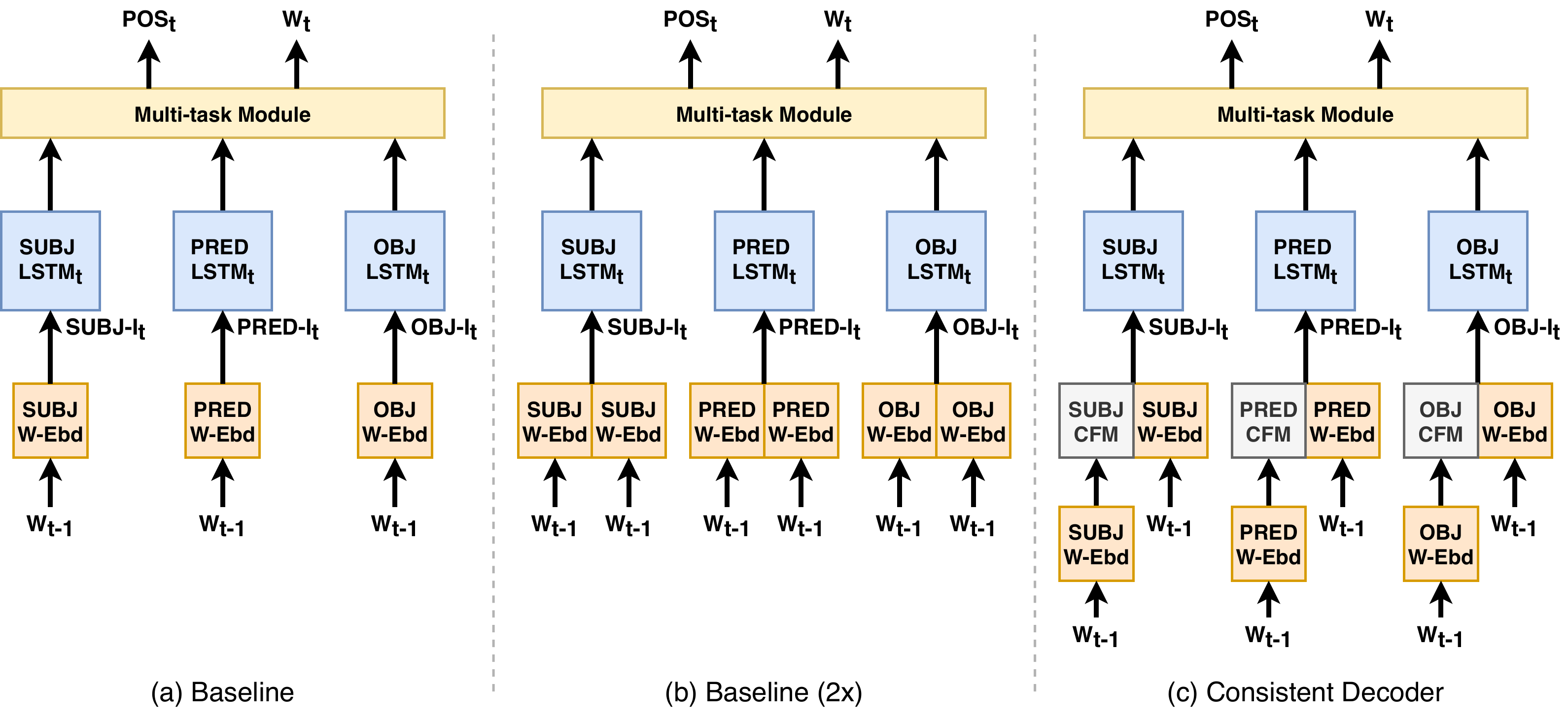}
  \caption{{\bf Captioning Experiment.} One recurrent unit at time step $t$ of the models. (a) Baseline, the MTTSNet~\cite{kim2019dense} model, where \{SUBJ, PRED, OBJ\} LSTMs are the triple-stream decoders. \{SUBJ, PRED, OBJ\} W-Ebds are the respective word embedding layers. $\text{POS}_t$ and $\text{W}_t$ are the POS tag and word output predicted at time step $t$. (b) Baseline (2x). (c) Our consistent decoder, where we have one consistent fusion mechanism (CFM) for every LSTM stream.}
  \label{fig:model_caption}
\end{figure}

\vspace{0.1in}
\noindent
\textbf{Models.} As in the mathematical sequence decoding section, we also have three models for this dataset: Baseline, Baseline (2x), and our Consistent Decoder.

\vspace{0.1in}
\noindent
\textbf{Baseline.} \cite{kim2019dense} proposes a multi-task triple-stream network (MTTSNet) for dense relational captioning. MTTSNet has three LSTM RNNs to generate outputs based on the subject region (SUBJ), object region (OBJ), and the union region (PRED) respectively. Then the outputs from the three RNNs are concatenated to predict a caption word and its part-of-speech ($<$POS$>$) tag ({\tt subject}, {\tt predicate}, or {\tt object}) in one time step via a multi-task module. We treat this as our baseline. One such recurrent unit is illustrated in Figure~\ref{fig:model_caption}, comprising of subject, object, union embedding layers, LSTM units, and a multi-task module.

\vspace{0.1in}
\noindent
\textbf{Baseline (2x).} To mimic the structure of our consistent decoder, which receives two inputs, as in the Baseline (2x) in Section~\ref{sec:math_decoding}, we also build another Baseline (2x) on the dense relational captioning dataset. One such recurrent unit is shown in Figure~\ref{fig:model_caption}. Built upon the baseline, the other input to the LSTM unit is the identical output word embedding from the decoder's own previous time step.
% of its own, same as the original input. 

\vspace{0.1in}
\noindent
\textbf{Consistent decoder.} Our consistent decoder structure is the same as the Baseline (2x), while the other input to the LSTM unit is the fused output word embedding from the previous time step of all the decoders. We apply our consistency fusion mechanism over the output word embeddings from all the decoders. We define that two captions are correlated if they involve a same proposed bounding box. We use this correlation information to build the adjacency matrix. One such recurrent unit is illustrated in Figure~\ref{fig:model_caption}. 
% for our consistent decoder. After decoding one word at every time step for one sequence decoder, we apply our proposed consistent fusion mechanism onto the outputs of all the sequence decoders. The outputs from the consistent fusion are treated as the other input to every sequence decoder at next time step.

%The RNN input dimensions of our consistent decoder and the baseline are different. To eliminate the performance difference caused by the different RNN input dimensions, we propose another baseline called baseline extended, where it has the same structure as the baseline but has an RNN input dimension 1024. The values of the first 512 dimension are zeros, and the remaining are the original RNN input. 

\vspace{0.1in}
\noindent
\textbf{Implementation details.}
Rather than training the whole detection and captioning framework end-to-end from scratch, we utilize the trained detection module in the original dense relational captioning model provided by~\cite{kim2019dense}\footnote{\url{https://github.com/Dong-JinKim/DenseRelationalCaptioning}}. We use the trained detection module to extract region proposal (bounding box) locations and the relevant features. Given an image, we run the trained detection module, and keep the most confident $100$ proposed regions. We then apply one round of non-maximum suppression (NMS) with intersection over union (IoU) threshold $0.3$ to reduce the number of overlapping region proposals. The remaining region proposals are treated as detected bounding boxes for an image. On average, in the validation set, there are $14.47$ detected bounding boxes per image. We fix the detected bounding box information, and train the captioning module from scratch. Training requires pairs of bounding boxes. We do a pairwise combination on the detected bounding boxes and keep those which have corresponding ground-truth relational captions for training, which results in an actual number of $70,302$ training images. A proposed bounding box pair has a ground-truth relational caption if the two proposed bounding boxes both have non-zero overlaps with their respective ground-truth bounding boxes.
% in the ground truth label. 
On average, in the validation set, there are $19.94$ bounding box pairs having ground-truth captions per image.     

\vspace{0.1in}
\noindent
\textbf{Network setting.} Same as in~\cite{kim2019dense}, we feed the visual features to the relevant LSTMs in the first time step, and a special start of sequence token ($<$SOS$>$) at the second time step for starting decoding captions. The visual features and word embedding dimensions are set to 512. The LSTM input dimension is set to 512 for Baseline, and 1024 for Baseline (2x) and our Consistent Decoder. In the first two time steps of the Baseline (2x) and the Consistent Decoder, the other input is a zero vector. For our consistent decoder, in the consistency fusion mechanism, we set the number of $\operatorname{AGGREGRATE}$ and $\operatorname{COMBINE}$ iterations ($K$) to 2. 
%\red{The detailed network structure?}
% the gated graph neural network with adaptive unsupervised attention onto it, and then treat it as another input to the sequence decoder at the next time step. 
% We utilize the ground-truth bounding boxes and relation pairs to train both baseline and our model. We apply a pretrained Faster R-CNN~\cite{ren2015faster} model to extract the bounding box features. The baseline decoder is the same as what is proposed in~\cite{kim2019dense}, the multi-task triple-stream network, 

\vspace{0.1in}
\noindent
\textbf{Training procedure.} We use SGD optimizer with batch size of $1$ to train both baselines and our consistent decoder. Initial learning rate is set to $10^{-3}$ and it is halved every 2 epochs. The minimum learning rate is capped at $10^{-6}$. Momentum is set to 0.98. %\red{We disregard the images which has more than $400$ matched relation pairs for training, which results in an actual number of ??? training images.}
Since we fix the bounding box locations and their feature representations, the models are trained with the captioning loss $L_{\mathrm{cap}}$ and the POS classification loss $L_{\mathrm{pos}}$: 
\begin{equation}
    L =  L_{\mathrm{cap}} + \lambda L_{\mathrm{pos}},
\end{equation}
where $\lambda$ is set to 0.1. $L_{\mathrm{cap}}$ and $L_{\mathrm{pos}}$ are both cross-entropy losses at every time step for word and POS classifications respectively. 

\vspace{0.1in}
\noindent
\textbf{Evaluation metrics.}
Our consistent decoder has the property that it can produce more consistent captions in terms of bounding box level descriptions. Besides only measuring caption correctness as in earlier works~\cite{johnson2016densecap,yang2017dense,yin2019context}, we propose a caption consistency evaluation to measure the consistency patterns exhibited in the generated captions. For completeness, we also report image level recall (Img-Lv. Recall) measurement~\cite{kim2019dense} to evaluate the word diversity of generated captions for a given image. However this measure is faulty; as it measures diversity of captions {\em within} an image. We argue one desires consistency {\em within} an image (an opposite of diversity) but diversity {\em across} images. To address this we introduce another bounding box level description diversity measurement. 

% However, this image level recall measurement is not meaningful for our setting. 

% Our consistent decoder will generate more consistent descriptions for a specific bounding box, which is desirable, thus will have less word diversity in terms of captions produced for one image. Therefore, we introduce another bounding box level description diversity measurement to evaluate the word diversity of descriptions for a specific bounding box in a specific relation.  
% Performance is in \%. 
% we use percentage as the unit.

\vspace{0.1in}
\noindent
\textbf{Caption correctness measurement.} Following~\cite{kim2019dense}, we use mean average precision (mAP) to measure both localization and language accuracy, and average METEOR score~\cite{banerjee2005meteor} to evaluate the caption correctness regardless of the bounding box location. When calculating mAP, the METEOR score thresholds for language are \{0, 0.05, 0.1, 0.15, 0.2, 0.25\}, and the IoU thresholds for bounding box localization are \{0.2, 0.3, 0.4, 0.5, 0.6\}. For the localization AP, both bounding boxes in the bounding box pair are measured with their respective ground-truths. Only the results with IoUs of both bounding boxes greater than the localization threshold are considered for the language thresholds.

\vspace{0.1in}
\noindent
\textbf{Caption consistency measurement.} We propose a measurement to evaluate the consistency of generated captions. In one image, if a bounding box appears in multiple relations, we calculate the average pair-wise BLEU-1 score~\cite{papineni2002bleu} among the generated descriptions for this specific bounding box. We utilize the predicted POS tag to extract the bounding box descriptions from a generated caption. We then take the mean of the averaged pairwise BLEU-1 scores over the whole test set to get the consistency score. Consistency score is in the range $[0, 1] \times 100$; higher is better. Values closer to $1$ imply more consistent results for one region.

\vspace{0.1in}
\noindent
\textbf{Bounding box level diversity measurement (BBox Div.).} % The aforementioned Img-Lv. Recall is problematic for our setting. Since our consistent decoder will generate more consistent captions, it will receive a lower value on this measurement. But if we use sampling for caption decoding, and generate captions for a given image multiple times, our consistent decoder should not negatively impact the word diversity in the multiple descriptions for a specific bounding box in a specific relation.
We argue that one desires a more consistent captioning when decoding multiple sequences involving the same region (bounding box). However, when we use sampling for caption decoding, generating captions for an image multiple times, while each individual run should be consistent, the results across runs should remain diverse. In other words a model should be able to choose among diverse references to use for a given region/object, but be consistent in its use of this reference in a given run. 
To capture this intuition we propose a word diversity measurement on the bounding box level to measure the diversity of the bounding box descriptions obtained with multiple runs of captioning for the same image. We calculate the pairwise BLEU-1 score~\cite{papineni2002bleu} among the multiple descriptions for a specific bounding box in a specific relation, and average over all the bounding boxes. This measure is treated as our diversity score on the bounding box level. The score is in the range $[0, 1] \times 100$. Lower score is better and indicates more diverse bounding box descriptions. 

\begin{figure}[t]
  \centering
  \includegraphics[width=\textwidth]{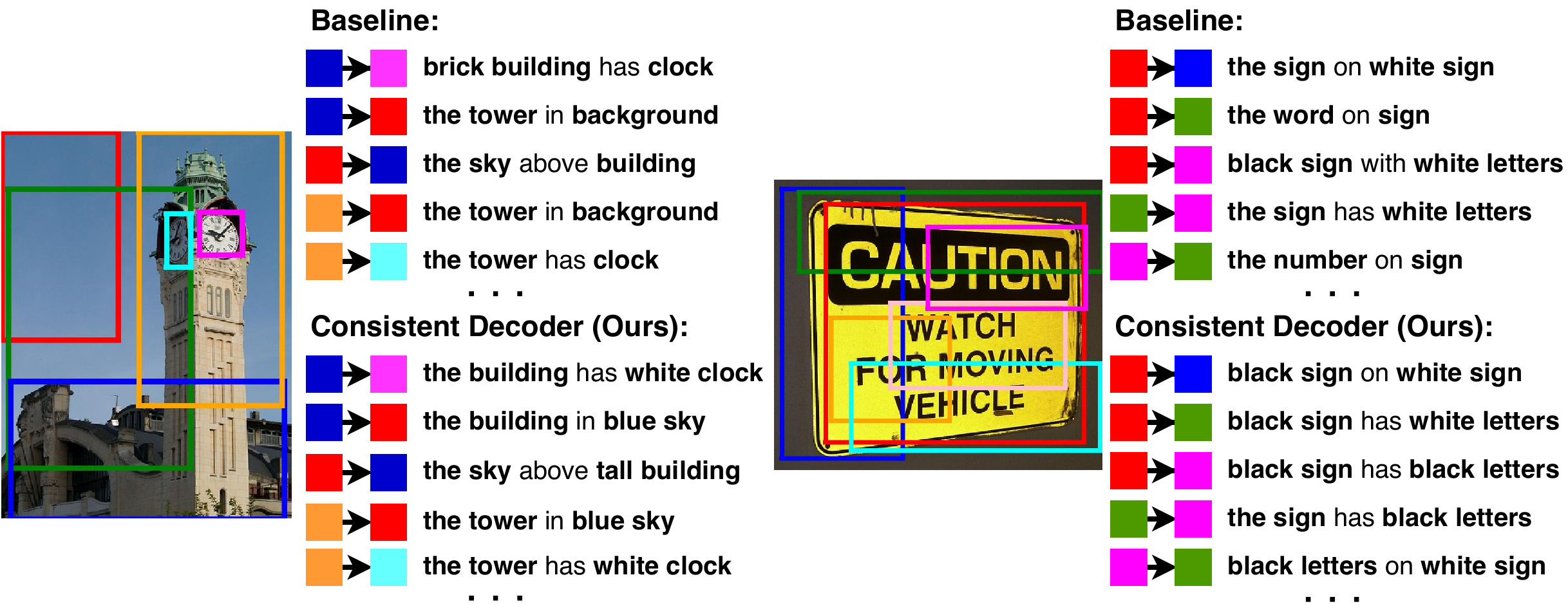}
  \caption{{\bf Qualitative results.} Example results for dense relational captioning 
  % Two qualitative results from the models 
  trained using the original label set. Given a pair of bounding boxes ({\tt subject} $\rightarrow$ {\tt object}), the generated captions are of form: ``\textbf{subject description} predicate \textbf{object description}".}
  \label{fig:results_qual}
\end{figure}

\vspace{0.1in}
\noindent
\textbf{Results.} % \footnote{We use percentage as the unit for reporting results here.}
During testing, we give the model the pairwise combination information of the detected bounding boxes for caption generation. When decoding captions, at every time step, we take the most probable word and POS tag for all the measurements except the BBox Div. measurement. The BBox Div. measurement requires sampling during caption decoding. On this metric, we caption $5$ times for a given image, which results in $5$ descriptions for a specific bounding box in a specific relation. We only measure this on the original label set because the ground-truth descriptions for a specific bounding box in the consistent label set are not diverse by design. Table~\ref{table:results_original} and Table~\ref{table:results_frequent} show the quantitative results on the original label set and the consistent label set respectively. For completeness, we report the direct results from~\cite{kim2019dense} (line~1 in Table~\ref{table:results_original}, MTTSNet-50 RP), where the model is trained end-to-end from scratch using the original label set, and evaluated using $50$ most confident region proposals before the second round of NMS. We also evaluate the same trained model %\footnote{\url{https://github.com/Dong-JinKim/DenseRelationalCaptioning}} 
using $100$ most confident region proposals (same as in our model). % , the same test setting as in our scenario. 
The results are listed in line~2 in Table~\ref{table:results_original} (MTTSNet-100 RP). Our baseline model has the same structure as the original relational captioning model~\cite{kim2019dense}. For our baseline, baseline (2x), and consistent decoder, only the decoding mechanism is trained on top of a fixed region detection backbone. Figure~\ref{fig:results_qual} shows two qualitative results from the baseline model and our consistent decoder trained with the original label set. 

% \caption{{\bf Test results on the original label set.} \red{We report direct results form~\cite{kim2019dense} in line 1, with 50 region proposals. We find that re-training~\cite{kim2019dense} with 100 region proposals further improves results (line 2). Baseline and Baseline (2x) are variants MTTSNet (100RP) where, similar to Consistent Decoder, only the decoding mechanism is trained on top of a fixed region detection backbone. Same training protocol, decoder capacity and structure are used for Baseline (2x) and  our Consistent Decoder. The only difference is that Consistent Decoder has proposed {\em consistency fusion mechanism}.}}

\begin{table}[t]
  \setlength{\tabcolsep}{3pt}
  \begin{center}
  \caption{{\bf Test results on the original label set.} Same training protocol, decoder capacity and structure are used for Baseline (2x) and our Consistent Decoder. The only difference is that Consistent Decoder has our proposed {\em consistency fusion mechanism}.}
  \label{table:results_original}
  \begin{tabular}{l|c|c|c|c|c}
  \hline
    Model & mAP & METEOR & Consistency & Img-Lv. Recall & BBox Div.\\
    \hline \hline
    MTTSNet-50 RP~\cite{kim2019dense} & 0.88 & 18.73 & - & 34.27 & - \\
    MTTSNet-100 RP~\cite{kim2019dense} & 1.29 & 20.26 & - & 55.68 & - \\
    Baseline & 1.93 & 22.86 & 33.08 & 71.32 & 17.26 \\
    Baseline (2x) & 1.94 & 22.86 & 33.27 & 71.35 & 17.07 \\
    Consistent Decoder & \textbf{2.04} & \textbf{23.32} & \textbf{36.43} & 69.60 & 17.50 \\
    \hline
  \end{tabular}
  \end{center}
  \vspace{-0.15in}
\end{table}

\begin{table}[t]
  \setlength{\tabcolsep}{3pt}
  \begin{center}
  \caption{{\bf Test results on the consistent label set.} Same training protocol, decoder capacity and structure are used for Baseline (2x) and our Consistent Decoder. The only difference is that Consistent Decoder has our proposed {\em consistency fusion mechanism}.}
  \label{table:results_frequent}
  \begin{tabular}{l|c|c|c|c}
  \hline
    Model & mAP & METEOR & Consistency & Img-Lv. Recall\\
    \hline \hline
    Baseline & 1.90 & 22.49 & 33.35 & 71.58 \\
    Baseline (2x) & 1.93 & 22.45 & 33.50 & 71.60 \\
    Consistent Decoder & \textbf{2.01} & \textbf{22.58} & \textbf{38.71} & 69.60 \\
    \hline
  \end{tabular}
  \end{center}
  \vspace{-0.15in}
\end{table}

\vspace{0.1in}
\noindent
\textbf{Result analysis.} Our consistent decoder not only performs better than baselines on caption correctness, but also generates more consistent captions. Compared to the results of the trained model provided by~\cite{kim2019dense}, our consistent decoder achieves new state-of-the-art results. 
%on the dense relational captioning task. 
On the original label set, the consistent decoder has $5.2\%$ relative improvement on mAP and $9.5\%$ relative improvement on the consistency measure. The improvements are due solely to 
consistency fusion mechanism. Comparing the difference on the BBox Div. score among the models, % and the difference on the Consistency score, 
our consistent decoder does generate bounding box level descriptions which have similar word level diversity with the baselines. The consistent label set, by design, has consistent ground-truth bounding box descriptions. As a result, our relative improvement in the consistency score on the consistent label 
set ($+15.6\%$ over the baseline) is greater than that on the original one ($+9.5\%$).

\vspace{0.1in}
\noindent
\textbf{Graph neural network and self-attention matters.} Our consistency fusion mechanism is implemented by gated GNN~\cite{li2015gated} with self-attention. % , a type of graph neural networks. 
To investigate how the number of $\operatorname{AGGREGRATE}$ and $\operatorname{COMBINE}$ iteration ($K$), in the gated GNN, will impact our captioning results, we conduct experiments with $K = 1, 2, \text{and}~3$. Also, to explore the usefulness of gated GNN and self-attention, we conduct experiments using model without gated GNN (Ablation 1); and model with gated GNN ($K = 1$) but having equally weighted attention (Ablation 2). For the model without Gated GNN, the fused output representation is simply an unweighted average of word embeddings from all directly related decoders. All experiments here are performed on the original label set. The results are illustrated in Table~\ref{table:results_original_k}. Gated graph neural network and self-attention do help generate better captioning performance. With the increasing of $K$, the captioning results are improved. When $K = 1$, our consistency fusion mechanism can learn a weighted average of output representations from all directly relevant decoders. As $K$ increases, information can be propagated among decoders that do not directly effect each other, but do so indirectly through other intermediate decoders. This appears to be an important property, particularly for improving consistency. We find that $K = 2$ is a reasonable compromise between quality of results and running time for the dense relational captioning task. 

% The intuition behind the increasing of captioning performance with the increasing of $K$ is that \red{the confidence of generating a correct word will be increased by knowing the output from other decoders. This confidence increment will be accumulated when $K > 1$. (How to explain this more properly?)} Since larger $K$ will result in longer network training time, judging the trade-off between captioning performance and network training speed, we choose $K = 2$ in all our consistent decoders for the dense relational captioning task.

\begin{table}[t]
  \setlength{\tabcolsep}{3pt}
  \begin{center}
  \caption{Test results on the original label set with different model settings.}
  \label{table:results_original_k}
  \begin{tabular}{l|c|c|c|c}
  \hline
    Model & mAP & METEOR & Consistency & Img-Lv. Recall \\
    \hline \hline
    Ablation 1 (No Gated GNN) & 2.01 & 23.17 & 35.29 & 70.65 \\
    Ablation 2 (Gated GNN Only) & 2.02 & 23.20 & 35.39 & 70.60 \\
    Consistent Decoder ($K = 1$) & 2.04 & 23.26 & 35.71 & 70.21 \\
    Consistent Decoder ($K = 2$) & 2.04 & \textbf{23.32} & 36.43 & 69.60 \\
    Consistent Decoder ($K = 3$) & \textbf{2.05} & 23.27 & \textbf{36.47} & 69.50 \\
    \hline
  \end{tabular}
  \end{center}
  \vspace{-0.15in}
\end{table}

% % \vspace{0.1in}
% % \noindent
% % \textbf{Consistent decoder preserves caption diversity.} The aforementioned image-level recall measurement measures the word diversity on a caption level. Since our consistent decoder will generate more consistent captions, it will receive a lower value on this measurement. 
% % However, if we use sampling for caption decoding, rather than taking the most probable word at every time step, and generate captions for a given image multiple times, our consistent decoder should not affect the word diversity in the multiple descriptions for a specific bounding box in a specific relation. 
% To validate this thought, we generate captions using sampling $5$ times using the model trained with the original label set. Then we measure the diversity of the descriptions generated for a specific bounding box in a specific relation. One such bounding box will have $5$ generated descriptions. 
% We calculate the pair-wise BLEU-1 score among the $5$ descriptions, and average over all the bounding boxes. We define this results as our diversity score for the bounding-box level diversity measurement. This score is in the range $[0, 1] \times 100$, the lower the score, the more diverse of the bounding box descriptions. 
% The diversity results are listed in the right-most column of Table~\ref{table:results_original}. Comparing the difference on the diversity score among the models and the difference on the consistency score, our consistent decoder do generate bounding box level descriptions which have similar word-level diversity with the baselines. 

\section{Conclusion}
We propose a consistent multiple sequence decoder, which can utilize the consistent pattern in the sequences to be decoded. With the core consistency fusion mechanism component, one decoder can utilize the outputs from all other decoders for better decoding. We show the advantage of our consistent decoder on two datasets, one synthetic dataset having two correlated mathematical sequences, and one real dense relational captioning dataset. Our consistent decoder achieves new state-of-the-art results on the dense relational captioning task.  

% \clearpage
\bibliographystyle{splncs04}
\bibliography{egbib}

\begin{thebibliography}{10}
\providecommand{\url}[1]{\texttt{#1}}
\providecommand{\urlprefix}{URL }
\providecommand{\doi}[1]{https://doi.org/#1}

\bibitem{alahi2016social}
Alahi, A., Goel, K., Ramanathan, V., Robicquet, A., Fei-Fei, L., Savarese, S.:
  Social lstm: Human trajectory prediction in crowded spaces. In: Proceedings
  of the IEEE conference on computer vision and pattern recognition. pp.
  961--971 (2016)

\bibitem{anderson2018bottom}
Anderson, P., He, X., Buehler, C., Teney, D., Johnson, M., Gould, S., Zhang,
  L.: Bottom-up and top-down attention for image captioning and visual question
  answering. In: Proceedings of the IEEE conference on computer vision and
  pattern recognition. pp. 6077--6086 (2018)

\bibitem{antol2015vqa}
Antol, S., Agrawal, A., Lu, J., Mitchell, M., Batra, D., Lawrence~Zitnick, C.,
  Parikh, D.: Vqa: Visual question answering. In: Proceedings of the IEEE
  international conference on computer vision. pp. 2425--2433 (2015)

\bibitem{banerjee2005meteor}
Banerjee, S., Lavie, A.: Meteor: An automatic metric for mt evaluation with
  improved correlation with human judgments. In: Proceedings of the acl
  workshop on intrinsic and extrinsic evaluation measures for machine
  translation and/or summarization. pp. 65--72 (2005)

\bibitem{cho2014properties}
Cho, K., Van~Merri{\"e}nboer, B., Bahdanau, D., Bengio, Y.: On the properties
  of neural machine translation: Encoder-decoder approaches. arXiv preprint
  arXiv:1409.1259  (2014)

\bibitem{devlin2018bert}
Devlin, J., Chang, M.W., Lee, K., Toutanova, K.: Bert: Pre-training of deep
  bidirectional transformers for language understanding. arXiv preprint
  arXiv:1810.04805  (2018)

\bibitem{felsen2017will}
Felsen, P., Agrawal, P., Malik, J.: What will happen next? forecasting player
  moves in sports videos. In: Proceedings of the IEEE International Conference
  on Computer Vision. pp. 3342--3351 (2017)

\bibitem{gao2017video}
Gao, L., Guo, Z., Zhang, H., Xu, X., Shen, H.T.: Video captioning with
  attention-based lstm and semantic consistency. IEEE Transactions on
  Multimedia  \textbf{19}(9),  2045--2055 (2017)

\bibitem{hochreiter1997long}
Hochreiter, S., Schmidhuber, J.: Long short-term memory. Neural computation
  \textbf{9}(8),  1735--1780 (1997)

\bibitem{huang2019adaptively}
Huang, L., Wang, W., Xia, Y., Chen, J.: Adaptively aligned image captioning via
  adaptive attention time. In: Advances in Neural Information Processing
  Systems. pp. 8940--8949 (2019)

\bibitem{johnson2016densecap}
Johnson, J., Karpathy, A., Fei-Fei, L.: Densecap: Fully convolutional
  localization networks for dense captioning. In: Proceedings of the IEEE
  Conference on Computer Vision and Pattern Recognition. pp. 4565--4574 (2016)

\bibitem{kim2019dense}
Kim, D.J., Choi, J., Oh, T.H., Kweon, I.S.: Dense relational captioning:
  Triple-stream networks for relationship-based captioning. In: Proceedings of
  the IEEE Conference on Computer Vision and Pattern Recognition. pp.
  6271--6280 (2019)

\bibitem{kipf2016semi}
Kipf, T.N., Welling, M.: Semi-supervised classification with graph
  convolutional networks. arXiv preprint arXiv:1609.02907  (2016)

\bibitem{knyazev2019understanding}
Knyazev, B., Taylor, G.W., Amer, M.: Understanding attention and generalization
  in graph neural networks. In: Advances in Neural Information Processing
  Systems. pp. 4204--4214 (2019)

\bibitem{kosaraju2019social}
Kosaraju, V., Sadeghian, A., Mart{\'\i}n-Mart{\'\i}n, R., Reid, I.,
  Rezatofighi, H., Savarese, S.: Social-bigat: Multimodal trajectory
  forecasting using bicycle-gan and graph attention networks. In: Advances in
  Neural Information Processing Systems. pp. 137--146 (2019)

\bibitem{krishna2017dense}
Krishna, R., Hata, K., Ren, F., Fei-Fei, L., Carlos~Niebles, J.:
  Dense-captioning events in videos. In: Proceedings of the IEEE international
  conference on computer vision. pp. 706--715 (2017)

\bibitem{krishna2017visual}
Krishna, R., Zhu, Y., Groth, O., Johnson, J., Hata, K., Kravitz, J., Chen, S.,
  Kalantidis, Y., Li, L.J., Shamma, D.A., et~al.: Visual genome: Connecting
  language and vision using crowdsourced dense image annotations. International
  Journal of Computer Vision  \textbf{123}(1),  32--73 (2017)

\bibitem{lee2017desire}
Lee, N., Choi, W., Vernaza, P., Choy, C.B., Torr, P.H., Chandraker, M.: Desire:
  Distant future prediction in dynamic scenes with interacting agents. In:
  Proceedings of the IEEE Conference on Computer Vision and Pattern
  Recognition. pp. 336--345 (2017)

\bibitem{li2019entangled}
Li, G., Zhu, L., Liu, P., Yang, Y.: Entangled transformer for image captioning.
  In: Proceedings of the IEEE International Conference on Computer Vision. pp.
  8928--8937 (2019)

\bibitem{li2018jointly}
Li, Y., Yao, T., Pan, Y., Chao, H., Mei, T.: Jointly localizing and describing
  events for dense video captioning. In: Proceedings of the IEEE Conference on
  Computer Vision and Pattern Recognition. pp. 7492--7500 (2018)

\bibitem{li2015gated}
Li, Y., Tarlow, D., Brockschmidt, M., Zemel, R.: Gated graph sequence neural
  networks. arXiv preprint arXiv:1511.05493  (2015)

\bibitem{liu2019naomi}
Liu, Y., Yu, R., Zheng, S., Zhan, E., Yue, Y.: Naomi: Non-autoregressive
  multiresolution sequence imputation. arXiv preprint arXiv:1901.10946  (2019)

\bibitem{lu2016hierarchical}
Lu, J., Yang, J., Batra, D., Parikh, D.: Hierarchical question-image
  co-attention for visual question answering. In: Advances in neural
  information processing systems. pp. 289--297 (2016)

\bibitem{maas2013rectifier}
Maas, A.L., Hannun, A.Y., Ng, A.Y.: Rectifier nonlinearities improve neural
  network acoustic models. In: Proc. icml. vol.~30, p.~3 (2013)

\bibitem{olivastri2019end}
Olivastri, S., Singh, G., Cuzzolin, F.: End-to-end video captioning. In:
  Proceedings of the IEEE International Conference on Computer Vision
  Workshops. pp.~0--0 (2019)

\bibitem{pan2017video}
Pan, Y., Yao, T., Li, H., Mei, T.: Video captioning with transferred semantic
  attributes. In: Proceedings of the IEEE conference on computer vision and
  pattern recognition. pp. 6504--6512 (2017)

\bibitem{papineni2002bleu}
Papineni, K., Roukos, S., Ward, T., Zhu, W.J.: Bleu: a method for automatic
  evaluation of machine translation. In: Proceedings of the 40th annual meeting
  on association for computational linguistics. pp. 311--318. Association for
  Computational Linguistics (2002)

\bibitem{park2018sequence}
Park, S.H., Kim, B., Kang, C.M., Chung, C.C., Choi, J.W.: Sequence-to-sequence
  prediction of vehicle trajectory via lstm encoder-decoder architecture. In:
  2018 IEEE Intelligent Vehicles Symposium (IV). pp. 1672--1678. IEEE (2018)

\bibitem{rahman2019watch}
Rahman, T., Xu, B., Sigal, L.: Watch, listen and tell: Multi-modal weakly
  supervised dense event captioning. In: Proceedings of the IEEE International
  Conference on Computer Vision. pp. 8908--8917 (2019)

\bibitem{rennie2017self}
Rennie, S.J., Marcheret, E., Mroueh, Y., Ross, J., Goel, V.: Self-critical
  sequence training for image captioning. In: Proceedings of the IEEE
  Conference on Computer Vision and Pattern Recognition. pp. 7008--7024 (2017)

\bibitem{rhinehart2019precog}
Rhinehart, N., McAllister, R., Kitani, K., Levine, S.: Precog: Prediction
  conditioned on goals in visual multi-agent settings. In: Proceedings of the
  IEEE International Conference on Computer Vision. pp. 2821--2830 (2019)

\bibitem{shen2017weakly}
Shen, Z., Li, J., Su, Z., Li, M., Chen, Y., Jiang, Y.G., Xue, X.: Weakly
  supervised dense video captioning. In: Proceedings of the IEEE Conference on
  Computer Vision and Pattern Recognition. pp. 1916--1924 (2017)

\bibitem{sun2019stochastic}
Sun, C., Karlsson, P., Wu, J., Tenenbaum, J.B., Murphy, K.: Stochastic
  prediction of multi-agent interactions from partial observations. arXiv
  preprint arXiv:1902.09641  (2019)

\bibitem{tang2019multiple}
Tang, C., Salakhutdinov, R.R.: Multiple futures prediction. In: Advances in
  Neural Information Processing Systems. pp. 15398--15408 (2019)

\bibitem{velivckovic2017graph}
Veli{\v{c}}kovi{\'c}, P., Cucurull, G., Casanova, A., Romero, A., Lio, P.,
  Bengio, Y.: Graph attention networks. arXiv preprint arXiv:1710.10903  (2017)

\bibitem{vered2019joint}
Vered, G., Oren, G., Atzmon, Y., Chechik, G.: Joint optimization for
  cooperative image captioning. In: Proceedings of the IEEE International
  Conference on Computer Vision. pp. 8898--8907 (2019)

\bibitem{xu2018powerful}
Xu, K., Hu, W., Leskovec, J., Jegelka, S.: How powerful are graph neural
  networks? arXiv preprint arXiv:1810.00826  (2018)

\bibitem{yang2017dense}
Yang, L., Tang, K., Yang, J., Li, L.J.: Dense captioning with joint inference
  and visual context. In: Proceedings of the IEEE Conference on Computer Vision
  and Pattern Recognition. pp. 2193--2202 (2017)

\bibitem{yeh2019diverse}
Yeh, R.A., Schwing, A.G., Huang, J., Murphy, K.: Diverse generation for
  multi-agent sports games. In: Proceedings of the IEEE Conference on Computer
  Vision and Pattern Recognition. pp. 4610--4619 (2019)

\bibitem{yin2019context}
Yin, G., Sheng, L., Liu, B., Yu, N., Wang, X., Shao, J.: Context and attribute
  grounded dense captioning. In: Proceedings of the IEEE Conference on Computer
  Vision and Pattern Recognition. pp. 6241--6250 (2019)

\end{thebibliography}

\clearpage
\section{More Dense Relational Captioning Results}
\subsection{More Qualitative Results}
\begin{figure}[t]
  \centering
  \includegraphics[width=\textwidth]{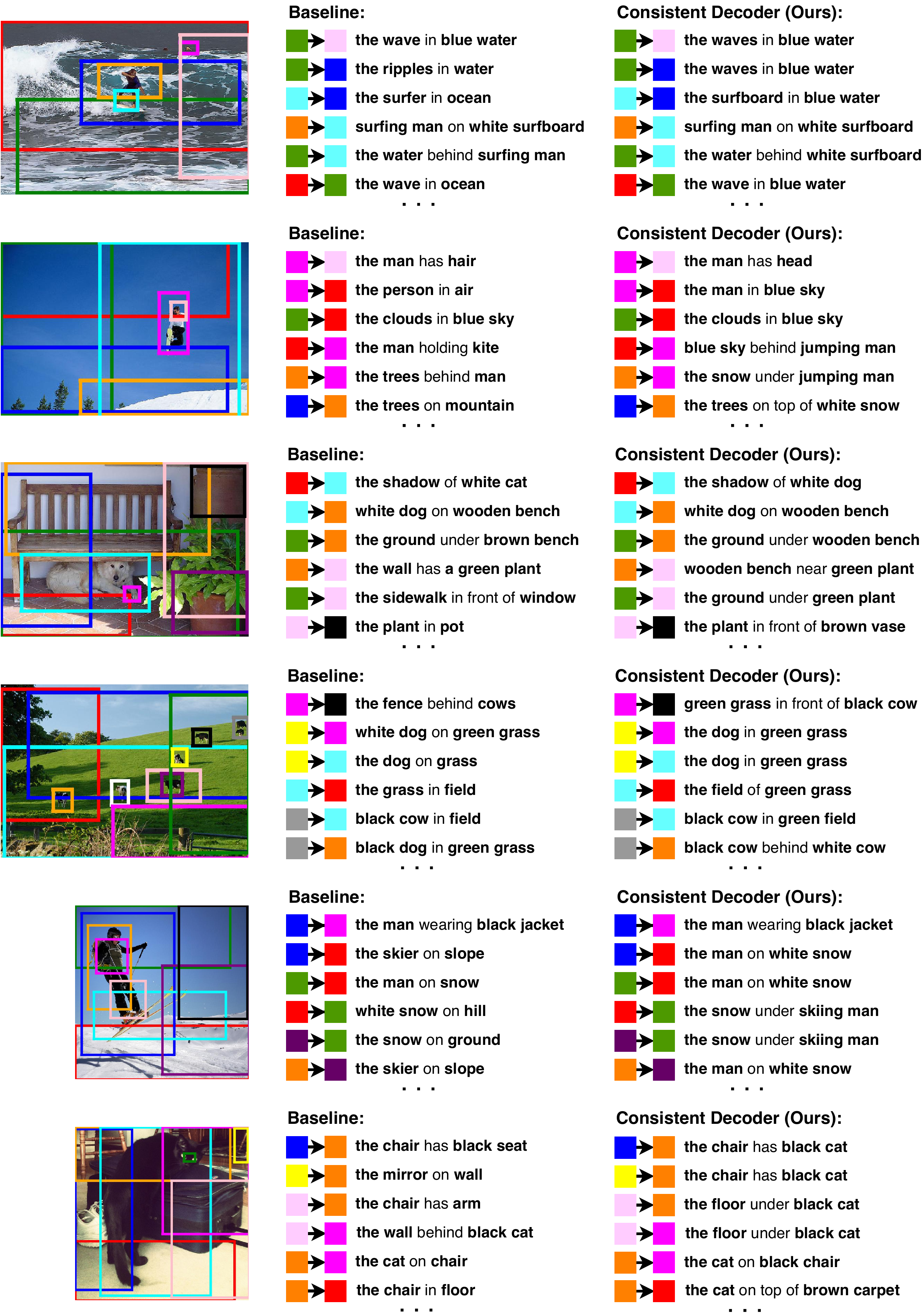}
  \caption{Qualitative results from the models trained with the original label set. Given a pair of bounding boxes ({\tt subject} $\rightarrow$ {\tt object}), the generated captions are of form: ``\textbf{subject description} predicate \textbf{object description}". Some bounding boxes are omitted for a clear visualization.}
  \label{fig:quali1}
\end{figure}

\begin{figure}[t]
  \centering
  \includegraphics[width=\textwidth]{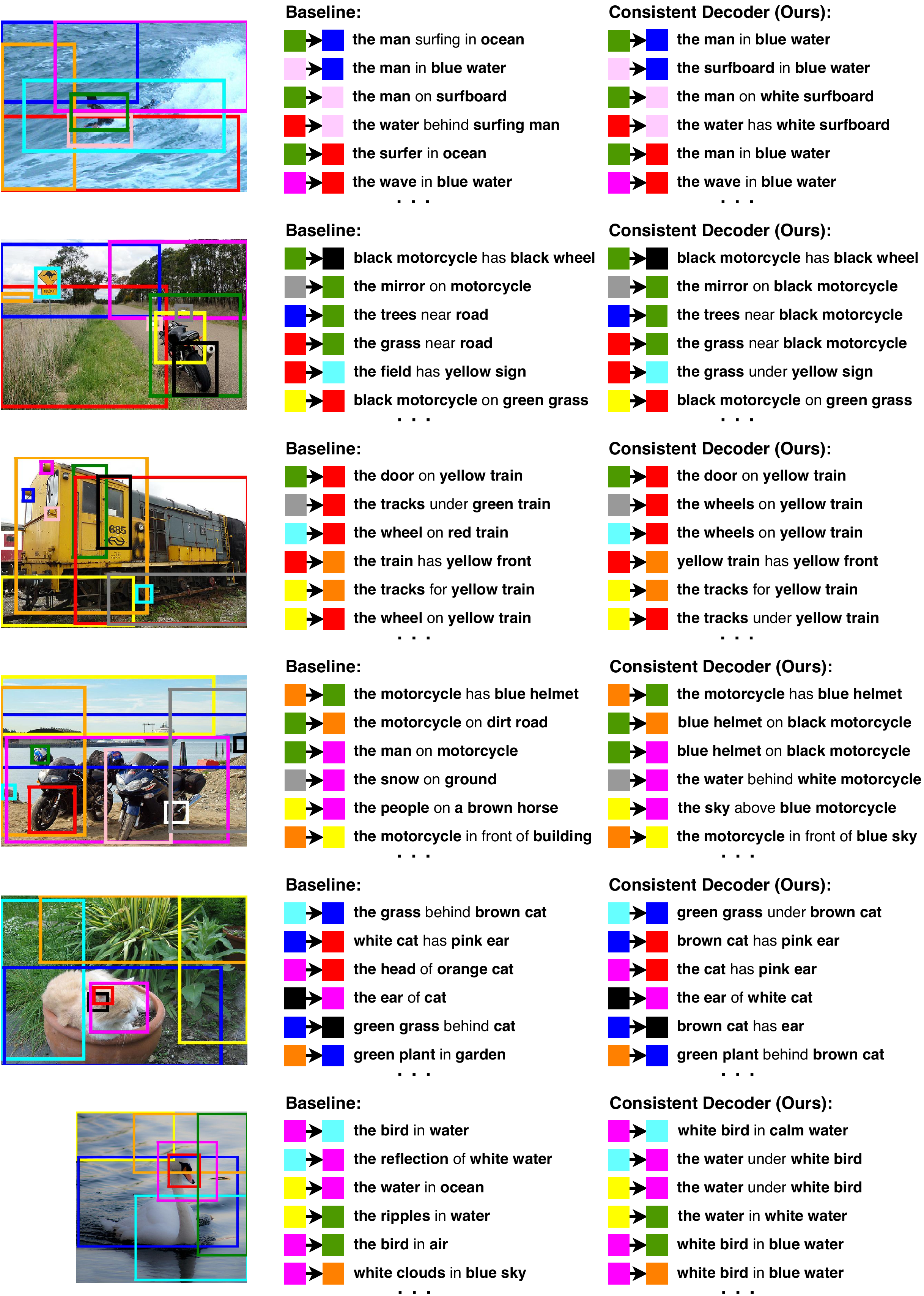}
  \caption{Qualitative results from the models trained with the original label set. Given a pair of bounding boxes ({\tt subject} $\rightarrow$ {\tt object}), the generated captions are of form: ``\textbf{subject description} predicate \textbf{object description}". Some bounding boxes are omitted for a clear visualization.}
  \label{fig:quali2}
\end{figure}

Figure~\ref{fig:quali1} and Figure~\ref{fig:quali2} show more qualitative results from the baseline model and our consistent decoder trained with the original label set. Captions are generated using the most probable word at every time step. 

\subsection{Diverse but Self-consistent Results}
\begin{figure}[t]
  \centering
  \includegraphics[width=\textwidth]{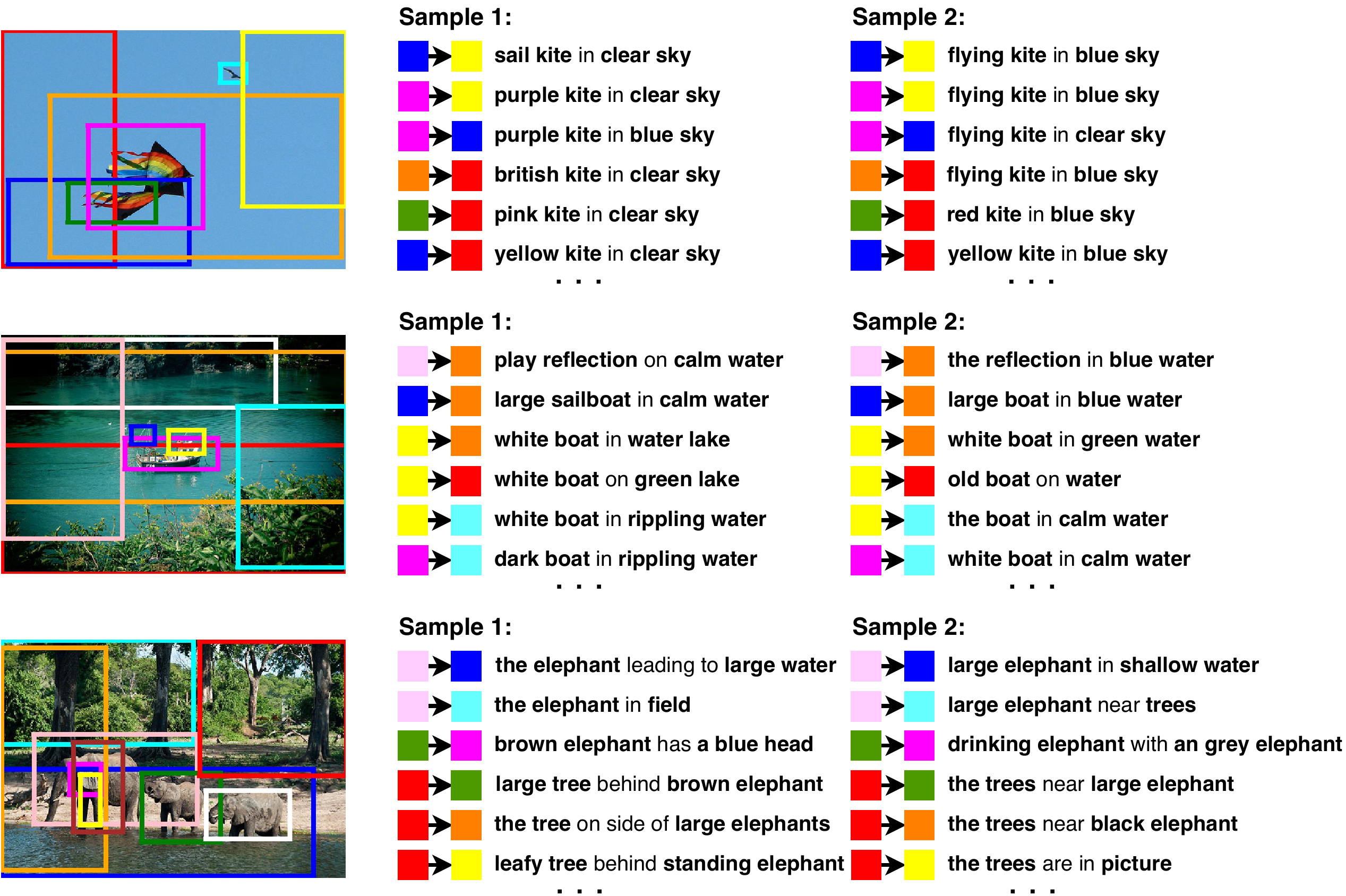}
  \caption{Qualitative diverse but self-consistent results from our consistent decoder trained with the original label set. Captions are generated using {\em sampling}.  Given a pair of bounding boxes ({\tt subject} $\rightarrow$ {\tt object}), the generated captions are of form: ``\textbf{subject description} predicate \textbf{object description}". Some bounding boxes are omitted for a clear visualization.}
  \label{fig:quali_diverse}
\end{figure}

Given an image, if we caption it multiple times using {\em sampling}, rather than taking the most probable word at every time step, with our consistent decoder, we should get multiple diverse but self-consistent caption sets for the image. Captions for one image should be diverse across multiple runs of captioning, but consistent in a single run. Figure~\ref{fig:quali_diverse} illustrates some such results from our consistent decoder trained with the original label set, where every image is paired with two sets of sampled captions. Note that when captioning with sampling, consistent with other image captioning approaches, due to inherent stochasticity, the individual caption quality is generally lower than that when captioning using the most probable word at each time step.

\end{document}